\documentclass[journal]{IEEEtran}
\ifCLASSINFOpdf
\else
\fi
\usepackage{times}
\usepackage{epsfig}
\usepackage{graphicx}
\usepackage{amsmath}
\usepackage{amssymb}
\usepackage{subfigure}
\usepackage{multirow}
\usepackage{color}
\usepackage[ruled]{algorithm2e}
\usepackage{ragged2e}
\usepackage{array}

\begin{document}
%
\title{Zero-Shot Learning via Latent Space Encoding}
%
%
%

\author{Yunlong~Yu$^\ddag$,~
        Zhong Ji*,~~\IEEEmembership{Member,~IEEE,}~
        Jichang Guo,
        and~Zhongfei~(Mark)~Zhang

}


\maketitle

\begin{abstract}
Zero-Shot Learning (ZSL) is typically achieved by resorting to a class semantic embedding space to transfer the knowledge from the seen classes to unseen ones. Capturing the common semantic characteristics between the visual modality and the class semantic modality (e.g., attributes or word vector) is a key to the success of ZSL. In this paper, we propose a novel encoder-decoder approach, namely Latent Space Encoding (LSE), to connect the semantic relations of different modalities. Instead of requiring a projection function to transfer information across different modalities like most previous work, LSE performs the interactions of different modalities via a feature aware latent space, which is learned in an implicit way. Specifically, different modalities are modeled separately but optimized jointly. For each modality, an encoder-decoder framework is performed to learn a feature aware latent space via jointly maximizing the recoverability of the original space from the latent space and the predictability of the latent space from the original space. To relate different modalities together, their features referring to the same concept are enforced to share the same latent codings. In this way, the common semantic characteristics of different modalities are generalized with the latent representations. Another property of the proposed approach is that it is easily extended to more modalities. Extensive experimental results on four benchmark datasets (AwA, CUB, aPY, and ImageNet) clearly demonstrate the superiority of the proposed approach on several ZSL tasks, including traditional ZSL, generalized ZSL, and zero-shot retrieval (ZSR).
\end{abstract}


\IEEEpeerreviewmaketitle

\section{Introduction}

\IEEEPARstart{A}{lthough} the success of Convolutional Neural Network (CNN) \cite{zhang17cyb,pang17cyb,majumder16cyb,Simonyan15iclr,szegedy2015cvpr,He2016cvpr,cheng18tgrs} greatly enhances the performance of object classification, many existing models are based on supervised learning and require labour-intensive work to collect a large number of annotated instances for each involved class. Besides, the models have to be retrained again if new classes are added to the classification system, which brings a huge computational burden. These issues severely limit the scalability of the conventional classification models.

Zero-Shot Learning (ZSL) \cite{Fu14pami,Fu15pami,guo17tip,yang17cvpr,Akata15cvpr,li2015iccv,Chao2016eccv} enables a classification system to classify instances from unseen categories in which no data are available for training, and attracts a large amount attention in recent years. It is typically achieved by transferring the knowledge from abundantly labeled seen classes to no labeled unseen classes via a class semantic embedding space, where the names of both the seen and unseen classes are embedded as vectors called class prototypes. Such a space can be human-defined attribute space \cite{farhadi2009cvpr,lampert13pami,Hwang2011cvpr} spanned by the pre-defined attribute ontology, or word vector space spanned by a large text corpus based on an unsupervised language processing technology \cite{Mikolov13nips,Pennington14EMNLP}. To this end, the semantic relationships between both the seen and unseen classes can be directly measured with the distances of the class prototypes in the class semantic embedding space.

In general, the performances of ZSL rely on the following three aspects: i) the representations of the visual instances; ii) the semantic representations of both the seen and unseen classes; and iii) the interactions between the visual instances and the class prototypes. On the one hand, the representations of visual instances are obtained with the off-the-shelf Convolutional Neural Networks (CNN), such as VGG \cite{Simonyan15iclr}, GoogleNet \cite{szegedy2015cvpr}, and ResNet \cite{He2016cvpr}. On the other hand, the class semantic embeddings are as important as the visual representations. The existing class prototypes are also collected in advance. In this way, the instance visual representations and the class semantic representations are obtained dependently. With the availability of the visual features and semantic class prototypes, the existing ZSL approaches mainly focus on learning a generalized interactional model to connect the visual space and the class semantic embedding space with the labeled seen classes only. ZSL is then achieved by resorting to the semantic distances between the unseen instances and the unseen class prototypes with the learned interactional model.

\begin{figure*}
\begin{center}
   \includegraphics[height=6.3cm,width=0.9\linewidth]{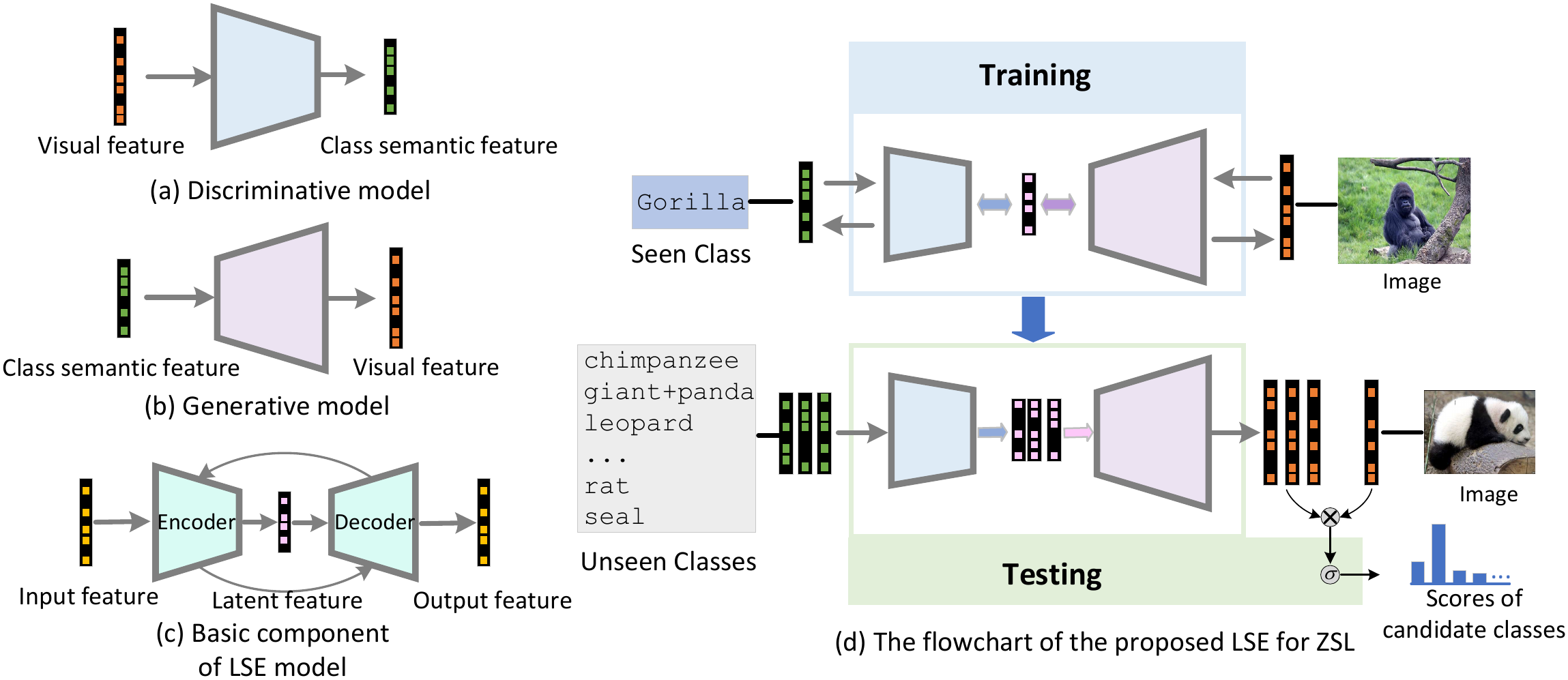}
\end{center}
   \caption{(a) and (b) are two existing standard ZSL models. Both models learn a projection function to connect different modalities. (c) illustrates the basic component of our LSE model. Each modality is fed into a constrained encoder-decoder framework to learn a latent space. (d) details the flowchart of LSE model for ZSL. Different modalities are enforced to share the same latent space to learn the common semantic characteristics. Semantic features of all candidate classes are then encoded in the visual space to classify test instances by comparing the compatibility scores between the test instances and all candidate classes.}
\label{fig:fig1}
\end{figure*}

The approaches of constructing the interactions between the visual space and class semantic embedding space can be divided into two categories: i) the label-embedding approaches (i.e., discriminative models), and ii) the visual instance generative approaches (i.e., generative models). Specifically, the label-embedding approaches \cite{lampert13pami,socher13nips} focuses on abstracting the high-level class semantic features from visual instances by learning a general function to project the visual features to the class semantic space. The testing unseen instances are then classified by matching the representations of the visual instances in the class semantic embedding space with the unseen class prototypes. On the other hand, the visual instance generative approaches \cite{Shigeto15ECML,lizhang17cvpr,Jiang17arxiv} learn an inversely projective function to generate the pseudo visual instances with the class semantic representations. In this way, the testing unseen instances are classified by resorting to the most similar pseudo visual instances of the unseen classes in the visual space. Experimental results show that the generative approaches perform better than label-embedding approaches since the latter is prone to suffering from hubness issue.

The existing approaches mostly require to learn an explicit projective function to relate different modalities. However, since the optimal projective function between two different spaces can be complicated and even indescribable, assuming that an explicit encoding function may not well model it. Besides, each modality has the distinctive characteristics despite of the common semantic information shared across different modalities, making the explicit encoding be easily spoiled. Assuming that different modalities referring to the same concept share some common semantic characteristics, we propose an encoder-decoder approach termed Latent Semantic Encoding (LSE) to explore the common semantic patterns across different modalities. Rather than learning a direct projection function, our LSE model uses matrix decomposition to expand a latent space from the input modality, which is an implicit process. It directly decomposes the input representation of each modality into a latent vector and an encoding matrix without any assumption of this encoding process. That is to say, the latent vector and the encoding matrix are to be learned. Compared to the discriminative and generative approaches that the input and output representations are required, this implicit encoding can reduce the risk of using an inappropriate projection function. In our work, the encoding matrix and decoding matrix are symmetric so that they can be modeled by the same set of parameters. The differences of LSE model and the current discriminative and generative approaches are illustrated in Fig. 1~(a)~(b)~(c).

To relate different modalities, we enforce different modalities share the same latent space spanned by the common semantic characteristics.
Such a constraint makes the common semantic characteristics be effectively explored via a common latent space. Besides, the proposed model can be easily integrated with more different modalities into the framework that makes use of complementary information to improve the performance even when further needed. Specific to ZSL, with the learned model, the visual embeddings of classes' semantic representations are constructed through the common latent space. Thus the test instances are classified according to the similarity scores of their visual vectors and the constructed visual representations. The flowchart of LSE model for ZSL is illustrated in Fig.~1~(d). 

In summary, our main contributions can be summarized into three folds:
\begin{enumerate}
  \item We introduce an encoder-decoder framework to exploit the intrinsic co-occurrence semantic patterns of different modalities. In this way, a better latent space for mitigating the distribution divergence across seen and unseen classes can be recovered.
  \item The symmetric constraint of the encoder-decoder framework ensures that the features are easily recovered via the other modalities, which captures the transferability and discriminability of the proposed approach.
  \item We also demonstrate that the proposed framework is suitable for multi-modality issue via exploring both the common and complementary information among different modalities. The experimental results show that finding an appropriate weight for each modality can yield an improved performance compared with that of any single modality.
\end{enumerate}


 We conduct extensive experiments on four datasets, i.e., Animal with Attribute (AwA) \cite{lampert13pami}, Caltech UCSD Birds (CUB) \cite{Catherine11} and aPY \cite{farhadi2009cvpr} attribute datasets, and ImageNet \cite{deng09cvpr}. The experimental results on traditional ZSL (TZSL), generalized ZSL (GZSL), and zero-shot retrieval (ZSR) tasks demonstrate that the proposed approach can not only transfer the source information to the target domain well but also preserve the discriminability between the seen classes and unseen ones.
\section{Related Work}
\begin{table}[ht]
\caption{\upshape The differences and relations of TZSL, GZSL, and ZSR.}
\begin{tabular}{|m{0.5cm}|p{0.4\columnwidth}|p{0.4\columnwidth}|}
 \hline
  Task & Differences \centering & \multicolumn{1}{c|}{Relations} \\
 \hline
  TZSL  & Test instances are assumed to be only from unseen classes. &\multirow{2}{*}\centering
  {The training process contains three steps: } \\
 \cline{1-2}
  GZSL & Test instances are assumed to be from both unseen as well as seen classes. & {image featurization, class semantic featurization, and train an interactional model to con-} \\
 \cline{1-2}
 ZSR & Retrieve instances according to class semantic information. &{nect visual and class semantic modalities.} \\
 \hline
\end{tabular}
\end{table}

Our work is related to several zero-shot learning scenarios, including TZSL, GZSL, and ZSR tasks. We review the differences and connections with respect to the related work separately.

\subsection{Traditional Zero-Shot Learning (TZSL).}
Inspired by the human-being's inferential ability that can recognize unseen categories according to the experiential knowledge about the seen categories and the descriptions of the unseen categories, TZSL is first attempted in \cite{Larochelle08aaai}, which introduces a model to generalize the unseen classes or tasks via their corresponding class descriptions. Motivated by this transferring mechanism, \cite{lampert13pami} represents each class with its corresponding class-level attributes and introduces two probabilistic models for TZSL. Considering that the collection of class-level attributes is a time-consuming work, \cite{socher13nips} and \cite{frome13nips} incorporate the natural language techniques into ZSL and use a high dimensional word vector to represent the name of each class. Likewise, \cite{Akata15cvpr} also represents the ontological relationships between different classes using the WordNet taxonomy. Once obtained the class semantic representations, the subsequently TZSL approaches mainly focus on learning the interactions between the visual modality and the class semantic modality. It is a cross-modality problem since the visual features and the class semantic features are located in different modalities. The existing approaches can be divided into three categories according to the direction of the mapping function between the visual space and the class embedding space. First, the simplest way is to learn a model to project the visual features to the class embedding space via Linear Regression \cite{Lazaridou14acl} or Neural Network \cite{socher13nips}. However, such a directional mapping easily suffers from the hubness issue \cite{marco2015hubness}, that is, the tendency of some unseen class prototypes (``hubs") appearing in the top neighbors of many test instances. To address this issue, Shigeto et al. \cite{Shigeto15ECML} propose to learn a reverse directional mapping function to project the class semantic embedding vectors to the visual space. Inspired by the cross-modality learning, reconstructing the interactions by learning a common latent space for both the visual and the class semantic embedding space is mostly focused on. By constructing a bilinear mapping, DeViSE \cite{frome13nips}, SJE \cite{Akata15cvpr}, ESZSL \cite{Romera15icml}, and JEDM \cite{yu2017trans} learn a translator function to measure the linking strength between the image visual features and the class semantic vectors.

\subsection{Generalized Zero-Shot Learning (GZSL).}
TZSL assumes that the testing instances are only classified into the candidate unseen classes. This scenario is unrealistic since the instances from seen classes have better chance to be tested in the real world. GZSL \cite{Chao2016eccv,fu16cvpr} is a more open setting that classifies the testing instances into both the seen and unseen classes. Compared with the TZSL, GZSL is a more challenging task. It requires not only to transfer the information from the source domain to the target domain but also to distinguish the seen classes and unseen ones. This is a dilemma since the effective transferability of the unseen classes relies on more related seen classes and however, more related seen classes dismiss the discriminability between the seen classes and unseen ones. In other words, most testing instances tend to be classified into the affinal seen classes rather than their groundtruth unseen ones. Although the existing TZSL approaches can be applied to GZSL directly, their classification performances are poor. Recently, a few approaches try to address this issue. For example, \cite{Chao2016eccv} proposes a simple approach to balance two conflicting forces: recognizing data from seen classes versus those from unseen ones. In order to improve the discriminant ability between the seen classes and unseen ones, \cite{fu16cvpr} proposes a maximum margin framework for semantic manifold based
recognition to ensure that the instances are projected closer to their corresponding class prototypes than to others (both the seen and unsee classes).


\subsection{Zero-Shot Retrieval (ZSR)}
Given a testing instance, zero-shot classification is to classify it into its most relevant candidate class. In contrast, the task of ZSR is an inverse-process that retrieves some images related to the specified attribute descriptions of unseen classes. In this way, ZSR can be seen as a special case of cross modality retrieval task. The performance of ZSR relies on two aspects: i) the consistency between visual intra-class representations and ii) the effective semantic alignments between the different modalities. Many existing ZSL approaches \cite{zhang15iccv, Zhang16cvpr, bucher2016eccv, xu2017cvpr} are extended to retrieval tasks. However, just like TZSL and GZSL, most existing approaches only retrieve the instances from the unseen set, and the generalized ZSR is still an open issue.

To this end, the common practical and inevitable challenge of ZSL (TZSL, GZSL, and ZSR) consists exclusively in preserving the semantic consistence between different modalities. Capturing the common semantic characteristics between different modalities is thus a key to the success of ZSL. The differences and relations of TZSL, GZSL, and ZSR are shown in TABLE \uppercase\expandafter{\romannumeral1}.

\section{Proposed Approach}

In this section, we first describe the proposed LSE approach and then apply it to address ZSL.

\subsection{Preliminaries}
Suppose that we have $D$ modalities, each of which consists of $N$ instances from $M$ different classes. Let $\{(\mathbf{x}_i^{(j)},y^{(j)})_{j=1}^{N}\}_{i=1}^{D}$ denote all training instances, where $\mathbf{x}_i^{(j)}\in\mathcal{X}_i\subset\mathbb{R}^{F_i}$ is the $j$-th instance of the $i$-th modality, $F_i$ is the dimensionality of $i$-th modality,  and $y^{(j)}\in\{1,2,...,M\}$ is the label of the corresponding instance. The pair $\mathbf{x}_i^{(j)}$ and $\mathbf{x}_k^{(j)}$ share the same class label $y^{(j)}$, which means that they represent the same concept. Denote $\mathcal{X}_i$ and $\mathcal{X}_j$ as a kind of visual modality and class semantic modality, the discriminative models learn a projection function $\mathcal{F}: \mathcal{X}_i\mapsto\mathcal{X}_j$ for predicting the class semantic representations based on the visual vector, while the generative models learn a reverse projection function. The project function $\mathcal{F}$ can be linear or nonlinear. These direct projection functions may not well model the semantic interactions across different modalities since the various structures and information. To alleviate this issue, we connect different modalities by exploring their common principle semantic characteristics with an encoder-decoder framework. For each modality, the encoder decomposes the input vector as a latent representation as well as a encoding matrix, while the decoder reconstructs the input with the latent representation and a decoding matrix. By jointly maximizing the recoverability of the latent space as well as the semantic consistency of different modalities makes the latent space feature-aware. Compared to the direct models that learn an explicit projection function, our approach can reduce the risk of learning an inappropriate function since the latent space is learned by semantic-driven. And thus the semantic relations of different modalities are probably revealed well.
\subsection{Latent Space Encoding}
 For each modality, LSE formulates the input as an encoder-decoder framework to encode a list of compact semantic patterns derived from input space with an implicit matrix decomposition. In the encoding process, the input matrix $\mathbf{X}_i$ is decomposed as the product of a code matrix $\mathbf{C}$ consisting of code vectors and a linear encoding matrix $\mathbf{U}_i^{T}$, i.e., $\mathbf{X}_i\sim{\mathbf{U}_i^{T}\mathbf{C}}$, where the encoding matrix and the code matrix are learned. This is an implicit encoding process. In the decoding process, the code matrix is decomposed into the product of the decoding matrix $\mathbf{V}_i$ and the original input matrix $\mathbf{X}_i$, i.e., $\mathbf{C}\sim\mathbf{V}_i\mathbf{X}_i$. This ensures that the learned code vectors are directly derived from the original input features. Generally, the effectiveness of the encoder-decoder framework depends upon both the representability of the latent space and the recoverability of the original input space.

To improve both the recoverability of the original input space and the representability of the latent space, the difference between the input matrix $\mathbf{X}_i$ and the recovered one using the latent code matrix $\mathbf{C}$ and the encoding matrix $\mathbf{U}_i^{T}$ should be minimized. Meanwhile, the difference between the latent code matrix $\mathbf{C}$ and the recovered one using the decoding matrix $\mathbf{V}_i$ and the original input matrix $\mathbf{X}_i$, is also expected to be minimized. Denoting the formula as $\Theta(\mathbf{X}_i,\mathbf{C},\mathbf{U}_i, \mathbf{V}_i)$, we thus have: \\
\begin{equation}\label{Equ:equ1}
\Theta(\mathbf{X}_i,\mathbf{C},\mathbf{U}_i)= (1-\lambda)\|\mathbf{X}_i-\mathbf{U}_i^{T}\mathbf{C}\|_F^2 + \lambda\|\mathbf{C}-\mathbf{V}_i\mathbf{X}_i\|_F^2,
\end{equation}
where $\|\cdot\|_F$ is the \emph{Frobenius norm} of a matrix, $\lambda$ is a parameter for balancing two items. The first item decomposes the original input matrix $\mathbf{X}_i$ into an encoding matrix $\mathbf{U}_i^{T}$ and a latent code matrix $\mathbf{C}$, which ensures that the latent features well capture the original visual features. The second item customizes the latent code matrix $\mathbf{C}$ to be derived from the original input matrix via a decoding matrix. It should be noted that the encoding matrix and the decoding matrix are symmetric so that they can be represented by the same parameters, i.e., $\mathbf{U}_i=\mathbf{V}_i$. Such a design makes not only the original vector highly recoverable but also Eq.~(\ref{Equ:equ1}) to have a closed optimization solution introduced below.

Given $\mathbf{C}$, the optimal $\mathbf{U}_i$ to minimize $\Theta(\mathbf{X}_i,\mathbf{C},\mathbf{U}_i)$ can be obtained as the following closed-form expression by setting its derivative with respect to $\mathbf{U}_i$ to 0,\\
\begin{equation}\label{Equ:equ2}
     (1-\lambda)\mathbf{C}\mathbf{C}^{T}\mathbf{U}_i+ \lambda\mathbf{U}_i\mathbf{X}_i\mathbf{X}_i^{T} = \mathbf{C}\mathbf{X}_i^{T}.
\end{equation}

To avoid redundant information in the latent space and enable the latent code vectors to encode the original input feature more compactly, we assume that the dimensional axes of $\mathbf{C}$ are uncorrelated and thus orthonormal, as shown in Eq.~(\ref{Equ:equ3}).
\begin{equation} \label{Equ:equ3}
    \mathbf{C}\mathbf{C}^{T} = \mathbf{I},
\end{equation}
where $\mathbf{I}$ is the identity matrix. Consequently, we obtain the optimal $\mathbf{U}_i$ with a closed-form expression:
\begin{equation} \label{Equ:equ4}
    \mathbf{U}_i = \mathbf{C}\mathbf{X}_i^{T}(\lambda\mathbf{X}_i\mathbf{X}_i^{T}+(1-\lambda)\mathbf{I})^{-1}.
\end{equation}

To this end, the decoding matrix $\mathbf{U}_i$ can be derived from the latent code matrix and the original input matrix. Thus, the task is to find an efficient code matrix. Substitute the optimal $\mathbf{U}_i$ to Eq.~(\ref{Equ:equ1}) leading to:
\begin{equation}\label{Equ:equ5}
\begin{aligned}
    &\Theta(\mathbf{X}_i,\mathbf{C},\mathbf{U}_i) = Tr[(1-\lambda)~\mathbf{X}_i^{T}\mathbf{X}_i + \lambda~\mathbf{C}^T\mathbf{C}]\\ &~~~~~~~~~~-Tr[\mathbf{C}\mathbf{X}_i^{T}(\lambda~\mathbf{X}_i\mathbf{X}_i^{T}+ (1-\lambda)~\mathbf{I})^{-1}\mathbf{X}_i\mathbf{C}^{T}],
\end{aligned}
\end{equation}
where $Tr[\cdot]$ denotes the trace of a matrix. With $Tr[(1-\lambda)\mathbf{X}_i^{T}\mathbf{X}_i+ \lambda\mathbf{C}^T\mathbf{C}]$ being a constant, minimizing Eq.~(\ref{Equ:equ5}) is equivalent to maximize $Tr[\mathbf{C}\mathbf{X}_i^{T}(\lambda \mathbf{X}_i\mathbf{X}_i^{T}+(1-\lambda)\mathbf{I})^{-1}\mathbf{X}_i\mathbf{C}^{T}]$, which can be seen as an expression of the recoverability of the visual space and the representability of the latent space, i.e., $\Phi(\mathbf{X}_i,\mathbf{C},\mathbf{U}_i)$. Consequently, we can derive the following formula.
\begin{equation}\label{Equ:equ6}
\begin{aligned}
    &\Phi(\mathbf{X}_i,\mathbf{C},\mathbf{U}_i) = Tr[\mathbf{C}\mathbf{X}_i^{T}(\lambda\mathbf{X}_i\mathbf{X}_i^{T}+(1-\lambda)\mathbf{I})^{-1}\mathbf{X}_i\mathbf{C}^{T}]\\
    &s.t.  ~~~~~~~~~~~~~~~\mathbf{C}\mathbf{C}^{T} = \mathbf{I}.
\end{aligned}
\end{equation}

By replacing $\mathbf{C}^{T}$ with $\mathbf{C}$, Eq.~(\ref{Equ:equ6}) is rewritten as:
\begin{equation}\label{Equ:equ7}
\begin{aligned}
    &\Phi(\mathbf{X}_i,\mathbf{C}) = Tr[\mathbf{C}^{T}\Delta_i\mathbf{C}],\\
    &s.t.  ~~~~~~~~~~~~~~~\mathbf{C}^{T}\mathbf{C} = \mathbf{I}.
\end{aligned}
\end{equation}
where $\Delta_i=\mathbf{X}_i^{T}(\lambda\mathbf{X}_i\mathbf{X}_i^{T}+(1-\lambda)\mathbf{I})^{-1}\mathbf{X}_i$.

In summary, the latent code matrix is learned in an implicit manner by balancing the predictability and the recoverability of the latent space, making the latent space feature-aware.

\subsection{Capture the intrinsic co-patterns across modalities}
Let $\mathbf{X}_i=[\mathbf{x}_{i}^{(1)},...,\mathbf{x}_{i}^{(N)}]$ be the input original matrix of the $i$-th modality. The above proposed model can encode the original matrix as a matrix embedded in a latent space. If the pair $\mathbf{x}_i^{k}$ and $\mathbf{x}_j^{k}$ from different modalities represent the same concept, their latent representations are strongly correlated. Here we set the correlated modalities to share the same latent code representations. To this end, the final objective function of the proposed LSE can be obtained as follows.
\begin{equation}\label{Equ:equ8}
\begin{aligned}
    &\Psi(\mathbf{X}_i,\mathbf{C}) = Tr[\mathbf{C}^{T}\sum_{i=1}^{D}\Delta_i\mathbf{C}],\\
    &s.t.  ~~~~~~~~~~~~~~~\mathbf{C}^{T}\mathbf{C} = \mathbf{I}.
\end{aligned}
\end{equation}
where $\Delta_i= \mathbf{X}_i^{T}(\lambda\mathbf{X}_i\mathbf{X}_i^{T}+(1-\lambda)\mathbf{I})^{-1}\mathbf{X}_i$. Using the Lagrange multipliers approach, each column $\mathbf{C}_{.,j}$ of the optimal $\mathbf{C}$ is obtained to satisfy the following condition:

\begin{equation}\label{Equ:equ9}
  (\sum_{i=1}^{D}\Delta_i)\mathbf{C}_{.,j} = \mu_j\mathbf{C}_{.,j}.
\end{equation}

It can be seen that the optimization for $\mathbf{C}$ can be transformed to an eigenvalue problem. The normalized eigenvectors of $\sum_{i=1}^{D}\Delta_i$ corresponding to the top $d$ largest eigenvalues form the optimal code matrix $\mathbf{C}$. $d$ is the dimensionality of the latent space. Consequently, the principal co-patterns of different modalities are revealed via the common latent matrix $\mathbf{C}$.

It should be noted that the orthogonal assumption of the columns of the latent matrix and symmetric constraints of encoder and decoding matrixes make the learned parameter an explicit solution. Without these assumptions, the learned parameters cannot be obtained directly. In general, from the perspective of the encoding process, the proposed model is implicit. However, from the perspective of the final solution, the model obtains an explicit matrix, it is explicit.

The computational complexity of $\Delta_i$ is $O(p^{3}+p^{2}N+N^2p)$, where $p$ is the dimensionality of the input feature matrix $\mathbf{X}_i$, and $N$ is the number of the input training instances. Since the dimensionality of the latent space $d$ is much smaller than $N$, the eigenvalue problem of Eq.~(\ref{Equ:equ9}) can be solved efficiently with iterative methods like Arnoldi iteration \cite{Lehoucq96maa}, of which the optimal computational complexity is about $O(d^2N)$. In this way, the overall computational complexity of the proposed approach is $O(p^{3}+p^{2}N+N^2p+d^2N)$.

\subsection{Discussions}
For each modality, the encoding matrix and decoding matrix can be symmetric or asymmetric. Compared to asymmetric constraint, the symmetric constraint brings some good properties for LSE:
\begin{enumerate}
  \item To obtain a closed-form expression. If the encoding matrix and decoding matrix are not symmetric, the objective function cannot be derived easily, which can be seen in Eq. (4), and the optimal optimized matrix cannot be obtained with a closed-form expression.
  \item To efficiently optimize the objective function and help the model to obtain global optima. The symmetric constraint enables the objective function to be transformed into an eigenvalue problem for efficient optimization.
  \item To reduce the optimized parameters. The symmetric constraint of encoding matrix and decoding matrix reduce the optimized parameters. Once we obtain one matrix, the other one is obtained correspondingly.
\end{enumerate}

\begin{algorithm}[h]
    \caption{The implementation of LSE for TZSL}
    \label{Algorithm 1}
    \KwIn{
    $\mathbf{X}_i\in\mathbb{R}^{p_i\times{N}}$: the feature matrix of the $i$-th modality, $i=\{1,2\}$, where $\mathbf{X}_1$ denotes the visual feature matrix and $\mathbf{X}_2$ denotes the class semantic feature matrix;\\
    ~~ $\lambda$: the balancing parameter;\\
    ~~ $d$: the dimensionality of the latent space;\\
    ~~ $\mathbf{x}_t$: the testing instance;\\
    ~~ $\mathbf{A}_u$: the semantic feature matrix of unseen classes.
   }
    \KwOut{The predicted class labels of unseen data.}
    \textbf{Training}:\\
    ~~ 1:~$\Delta_i=\mathbf{X}_i^{T}(\lambda\mathbf{X}_i\mathbf{X}_i^{T}+(1-\lambda)\mathbf{I})^{-1}\mathbf{X}_i$\\
    ~~ 2:~$\Omega= \sum_{i=1}^{D}\Delta_i$\\
    ~~ 3:~$\mathbf{C}$ = \textit{eigenvector}($\Omega$,~$d$)~\{eigenvectors corresponding to the top $d$ largest eigenvalues\}\\
    ~~ 4:~The decoding matrix:\\
    ~~~~~~  $\mathbf{U}_i = \mathbf{C}\mathbf{X}_i^{T}(\lambda\mathbf{X}_i\mathbf{X}_i^{T}+(1-\lambda)\mathbf{I})^{-1}$.\\
    ~~ 5:~The encoding matrix:  $\mathbf{U}_i^{T}$.\\
    \textbf{Testing}:\\
    ~~ 6:~Obtaining the latent representations of all the unseen classes with semantic features $\mathbf{A}_u$:\\
    ~~~~ $\mathbf{C}_u=\mathbf{U}_2\mathbf{A}_u$;\\
    ~~ 7:~Obtaining the visual representations of all the unseen classes with latent representations: \\
    ~~~~ $\widetilde{\mathbf{X}}_u=\mathbf{U}_1^{T}\mathbf{C}_u$;\\
    ~~ 8:~Obtaining the label of testing instance $\mathbf{x}_t$ with:\\
    ~~~~ $l(\mathbf{x}_t) = \arg\max_jcos(\mathbf{x}_t,\widetilde{\mathbf{x}}_j)$, where $\widetilde{\mathbf{x}}_j\in\widetilde{\mathbf{X}}_u$.
    \end{algorithm}

Most existing ZSL approaches learn a unidirectional projection function from one modality to the other one. These approaches may fail to uncover the distribution information of different modalities since
each class is only represented with a class-level semantic representation. However, LSE performs the multimodal interaction via a bidirectional way to improve the recoverability of the class semantic representation. The distribution of class semantic is derived from the visual distribution to the latent space based on the original class-level semantic representations, which is then used to reconstruct the visual distribution. In this way, LSE aligns the distribution of different modalities via the latent space, where the divergence and dissymmetry are adjusted.

\subsection{Apply LSE to ZSL}
Given a set of $N$ training instances from $C$ seen classes $\{\mathbf{x}_i,l_{i}\}_{i=1}^{N}$, where $\mathbf{x}_i\in\mathcal{X}$ and $l_{i}\in\mathcal{Y}=\{y_1,...,y_C\}$ are respectively the visual feature and label vector of the $i$-th instance. Zero-shot learning is to learn a classifier $f: \mathcal{X}\rightarrow\mathcal{Z}$ for a label set $\mathcal{Z}=\{z_1,...,z_L\}$ that is disjoint from $\mathcal{Y}$, i.e., $\mathcal{Y}\cap\mathcal{Z}={\O}$. In order to transfer the information from seen classes to unseen ones, each class $y\in\mathcal{Y}$ and $z\in\mathcal{Z}$ is associated with a semantic vector $\mathbf{a}\in\mathcal{A}$, e.g., attributes or word embedding.

ZSL can be seen as a special case of the proposed LSE approach, where the visual space is the first modality while the class semantic embedding space is the second modality. The visual modality and the class semantic modality are connected by learning a shared representation with Eq.~(\ref{Equ:equ8}). Once obtaining the optimal code matrix $\mathbf{C}$, the encoding and decoding matrices are easily derived with Eq.~(\ref{Equ:equ4}). In the testing stage, the unseen instances are classified by computing the similarity between the visual features and the unseen class semantic embedding vectors with the learned encoding and decoding matrices. In order to alleviate the influence of the hubness issue mentioned in \cite{marco2015hubness}, we perform ZSL in the visual space by encoding the class semantic vectors into the visual space:

\begin{equation}\label{Equ:equ10}
  l(\mathbf{x}_t) = \arg\max_jcos(\mathbf{x}_t,\widetilde{\mathbf{x}}_j),
\end{equation}
where $l(\mathbf{x}_t)$ returns the label of the test instance $\mathbf{x}_t$, $\widetilde{\mathbf{x}}_j=\mathbf{U}_1^T\mathbf{U}_2\mathbf{a}_j$ is the vector in the visual space projected by the $j$-th unseen class embedding vector $\mathbf{a}_j$, $cos(\mathbf{x}_t,\widetilde{\mathbf{x}}_j)$ is the compatibility score between the test instance and the class embedding vector. An illustration of the implementation of LSE for TZSL is shown in Algorithm 1.

For some cases, there are more than one type of class semantic embedding spaces available, each capturing an aspect of the structure of the class semantics. To explore the complementary information of different modalities, we can learn a better code matrix by combining them together. By learning the latent code matrix with Eq.~(\ref{Equ:equ8}), the encoder and decoder for each modality can be derived with Eq.~(\ref{Equ:equ4}) correspondingly. Consequently, we model the final prediction as

\begin{equation}\label{Equ:equ11}
  l(\mathbf{x}_t) = \arg\max_j\sum_k\alpha_kcos(\mathbf{x}_t,\widetilde{\mathbf{x}}_j^k),
\end{equation}
where $\widetilde{\mathbf{x}}_j^k$ is the vector in the visual space projected by the $j$-th unseen class embedding vector of modality $k$, $\alpha_k$ is the weight parameter for modality $k$. In our experiments, we perform a grid search over $\alpha_k$ on the unseen classes.

\section{Experiments}
In this section, we design extensive experiments to evaluate our proposed LSE approach. Firstly, we introduce the experimental setups, including the datasets, features, and evaluation metrics used in the experiments. Secondly, we provide TZSL, GZSL, and ZSR results on four benchmark datasets, respectively. Finally, some analysis of the proposed approach is further discussed.

\subsection{Experimental Setup}
\textbf{Datasets}. Three benchmark attribute datasets and a large-scale image dataset are used for our evaluations. (a) \textbf{Animal with Attributes (AwA)} dataset consists of 30,475 images from 50 animal classes. Each class is associated with 85 class-level attributes. We follow the same seen/unseen split as that in \cite{lampert13pami} for experiments. (b) \textbf{Caltech-UCSD Birds 2011 (CUB)} \cite{Catherine11} is a fine-grained dataset with 200 different bird classes, which consists of 11,788 images. Each class is annotated with 312 attributes. To facilitate direct comparison, we follow the split suggestion in \cite{Akata15cvpr}, of which 150 classes are used for training and the rest 50 classes for testing. (c) \textbf{aPascal-aYahoo} \cite{farhadi2009cvpr} is a combined dataset of aPascal and aYahoo, which contains 2,644 images from 32 classes. Each image is annotated with 64 binary attributes. To represent each class with an attribute vector, we average the attributes of the images in each class. In the experiments, the aPascal is used as the seen data, and aYahoo is used as the unseen data. (d) For \textbf{ImageNet} \cite{deng09cvpr}, we follow the same seen/unseen split as that in \cite{fu16cvpr}, where 1,000 classes from ILSVRC2012 are used for training, while 360 non-overlapped classes from ILSVRC2010 are used for testing. The details of these four datasets are listed in TABLE \uppercase\expandafter{\romannumeral2}.

\begin{table}
    \label{Table.5}
    \caption{\upshape The statistics of the four datasets used in the experiments. $\mathcal{A}$ and $\mathcal{W}$ are short for attribute space and word vector space, respectively.}
    \centering
    \begin{tabular}{|c|c|c|c|c|c|c|}
    \hline
    \multirow{2}{*}{Dataset} &\multicolumn{2}{c|}{SS} &\multicolumn{2}{c|}{Training} &\multicolumn{2}{c|}{Testing}\\
    \cline{2-7}
    & $\mathcal{A}$ & $\mathcal{W}$ &Images &Classes & Images &Classes\\
    \hline
    AwA & 85  &100 &24,295 &40  & 6,180 & 10 \\
    CUB & 312 &400 &8,855  &150 & 2,933 & 50 \\
    aPY & 64  & -  &12,695 &20  & 2,644 & 12 \\
    ImageNet &- &1,000 & 200,000 & 1,000 & 54,000 & 360\\
    \hline
    \end{tabular}
\end{table}

\textbf{Semantic embedding space}. For AwA and CUB datasets, both the attribute space and word vector space are used as the semantic embedding space. For an easy comparison with the existing approaches, we train a word2vector model \cite{Mikolov13nips} on a corpus of 4.6M Wikipedia documents to obtain the 100-dimensional vector for each AwA class name and 400-dimensional vector for each CUB class name. For aPY dataset, only the attribute space is used since few approaches are evaluated with word vector on it. For ImageNet dataset, 1,000-dimensional word vector is used to represent each class name.

\textbf{Visual feature space}. In order to better compare with the existing approaches, we use the deep features extracted from popular CNN architecture. For a fair comparison, two types of deep features: 4,096-dim VGG \cite{Simonyan15iclr} features and 1,024-dim GoogleNet \cite{szegedy2015cvpr} features are used for the three benchmark attribute datasets. Those features are available from \cite{zhang15iccv} and \cite{Changpinyo2016cvpr}, respectively. For the ImageNet dataset, we use the 1024-dim GoogleNet features provided in \cite{Kodirov17cvpr}.

\textbf{Evaluation metric}.
Following the traditional supervised classification, many ZSL approaches \cite{zhang15iccv,Kodirov17cvpr,xu2017cvpr} are evaluated with Per-image accuracy (PI), which focuses on classifying if the predicted label is the correct class label for each test instance. However, this criterion may encourage biased prediction in densely populated classes. Thus, the Per-class accuracy (PC) \cite{lampert13pami,yu2017trans,Changpinyo2016cvpr} is commonly used for ZSL. In our experiment, PC is adopted to evaluate ZSL and GZSL performances. For ZSR, mean average precision (mAP) \cite{zhang15iccv,bucher2016eccv,xu2017cvpr} is used to measure the performance.

\textbf{Implementation details}. Our LSE approach has two parameters to investigate, the balance parameter $\lambda$ and the dimensionality of latent space $d$. As in \cite{Zhang16cvpr}, their values are set by class-wise cross-validation using the training data. It should be noted that the dimensionality of the latent space is always smaller than that of the input space. All the experiments are conducted on a computer which has 4-core 3.3GHz CPUs with 24GB RAM.

\subsection{TZSL results}
\subsubsection{TZSL results with attribute}
\begin{table}\label{tab:tab1}
 \caption{\upshape{Comparison to the existing TZSL approaches in terms of classification accuracy (\%) on three datasets with attributes. Two types of deep features (VGG and GoogleNet) are used. $\mathcal{V}$ and $\mathcal{G}$ are short for VGG and GoogleNet features, respectively. `\S' ~indicates the methods with which the classification performances are obtained by ourselves. For each dataset, the best one with VGG features is marked with underline and the best one with GoogleNet features is marked in bold.}}

\begin{center}
\begin{tabular}{|l|c|c|c|c|}
\hline
Method & $\mathcal{F}$ &AwA & CUB & aPY\\
\hline\hline
DAP \cite{lampert13pami}   & $\mathcal{G}$ & 60.1 & 36.7 & 35.5 \\
RRZSL \cite{Shigeto15ECML} & $\mathcal{G}$ & 66.4 & 45.4 & 38.8 \\
ESZSL \cite{Romera15icml}\S & $\mathcal{G}$ & 76.8 & 49.1 & 47.3 \\
SAE \cite{Kodirov17cvpr}\S  & $\mathcal{G}$ & 81.4 & 46.2 & 41.3\\
SSE \cite{zhang15iccv}    & $\mathcal{V}$ & 76.3 & 30.4 & 46.2  \\
JLSE \cite{Zhang16cvpr}     & $\mathcal{V}$ & 80.5 & 41.8 & 50.4  \\
MLZSL \cite{bucher2016eccv} & $\mathcal{V}$ & 77.3 & 43.3 & \underline{53.2}  \\
MFMR \cite{xu2017cvpr}      & $\mathcal{V}$/$\mathcal{G}$ & 79.8/76.6 &47.7/46.2 & 48.2/46.4\\
SynC$^{struct}$ \cite{Changpinyo2016cvpr}   & $\mathcal{V}$/$\mathcal{G}$ & 78.6/73.4 &50.3/\textbf{54.4} & 48.9/44.2\\
\hline
LSE & $\mathcal{V}$/$\mathcal{G}$ &\underline{81.9}/\textbf{81.6} & \underline{55.4}/53.2 & 47.6/\textbf{53.9}\\
\hline
\end{tabular}
\end{center}
\end{table}

In order to evaluate the effectiveness of the proposed approach, nine state-of-the-art ZSL approaches are selected for comparison: 1) DAP \cite{lampert13pami}, RRZSL \cite{Shigeto15ECML}, ESZSL \cite{Romera15icml}, and SAE \cite{Kodirov17cvpr} are compared using the GoogleNet features; 2) SSE \cite{zhang15iccv}, JLSE \cite{Zhang16cvpr}, and MLZSL \cite{bucher2016eccv} are compared using the VGG features; 3) MFMR \cite{xu2017cvpr} and SynC$^{struct}$ \cite{Changpinyo2016cvpr} are compared using both the GoogleNet and the VGG features. The performance results of the selected approaches are all from the original papers except for ESZSL \cite{Romera15icml} and SAE \cite{Kodirov17cvpr}. ESZSL \cite{Romera15icml} and SAE \cite{Kodirov17cvpr} are fine tuned by ourselves using the codes released by the authors; the hyperparameters of both models are selected from \{0.01,0.1,1,10,100\}.

TABLE \uppercase\expandafter{\romannumeral3} presents the classification accuracy of our approach and the nine competitive baselines with attributes. Generally, our approach achieves the state-of-the-art performance on three benchmark datasets. Specifically, it outperforms all the competitors on AwA dataset, which has a 1.4\% and 0.2\% improvements over the closest VGG-based competitor (i.e., JLSE) and the closest GoogleNet-based competitor (i.e., SAE), respectively. In terms of the CUB dataset, the relative accuracy gain of LSE over SynC$^{struct}$ \cite{Changpinyo2016cvpr}, i.e., the second best approach, is 5.1\% with the VGG features. For aPY dataset, LSE also beats all the competitors with a large margin using the GoogleNet features. We also observe that different CNN features (i.e., VGG and GoogleNet) have slight different performances for the same approach, and the same visual feature may perform better in some approaches but worse in some other approaches. We argue that these are reasonable phenomena. Although both VGG and GoogleNet features are popular CNN-based visual features, their different network structures lead them to have similar but different performances for different approaches as well as on different datasets. From Table III, we can find that for most competitors, VGG features yield better performances than GoogleNet features on AwA and aPY datasets but worse on CUB dataset. It indicates that both two deep features have advantages relying on the datasets.

In addition, compared with the approaches that mostly focus on learning an explicit mapping function \cite{Shigeto15ECML,Romera15icml,zhang15iccv}, the proposed approach achieves obvious improvements on three datasets, showing that the effectiveness and the superiority of the our implicit encoding and the balance between the encoding and decoding processes.

\subsubsection{ZSL results on ImageNet dataset}

 Five state-of-the-art competitors are selected for the ImageNet dataset. Among them, DeViSE \cite{frome13nips} is an end-to-end deep embedding framework to connect the visual features and the word vectors via a common compatible matrix. By applying the embedding representations of the visual features with DeViSE, AMP \cite{fu15cvpr} constructs a class prototype graph to measure the similarity between the visual embedding representations and the class prototypes. ConSE \cite{Norouzi14iclr} learns an n-way probabilistic classifier for the seen classes and infers the unseen classifiers via the semantic relationships between the seen and unseen classes. SS-Voc \cite{fu16cvpr} and SAE \cite{Kodirov17cvpr} are two embedding approaches, in which SS-Voc \cite{fu16cvpr} improves classification performance by utilizing vocabulary over unsupervised items to train the model and SAE \cite{Kodirov17cvpr} adds a reconstructed term to encourage learning a more generalized model. The comparison results are demonstrated in TABLE \uppercase\expandafter{\romannumeral4}.

For a fair comparison with the alternatives, we use Top@1 and Top@5 classification accuracies to evaluate the approaches. From the comparison results, we can find that the proposed LSE obtains the superior performance on ImageNet dataset. Specifically, it outperforms the closest competitor over 1.8\% with top@5. This is impressive since the amount of testing instances is large.

\begin{table}\label{tab:tab1}
\caption{\upshape{The classification performance (\%) of different TZSL approaches on ImageNet dataset.}}
\begin{center}
\begin{tabular}{|l|c|c|}
\hline
Method  & Top@1 &Top@5\\
\hline\hline
DeViSE \cite{frome13nips} &5.2 & 12.8 \\
AMP \cite{fu15cvpr}       &6.1 & 13.1 \\
ConSE \cite{Norouzi14iclr} & 7.8 &15.5 \\
SS-Voc \cite{fu16cvpr}    & 9.5 &16.8 \\
SAE \cite{Kodirov17cvpr}\S &12.1 & 25.6 \\
LSE       & \textbf{12.4} & \textbf{27.4} \\
\hline
\end{tabular}
\end{center}
\end{table}

\subsubsection{ZSL results with multimodal features}

\begin{table}\label{tab:tab2}
\caption{\upshape{Comparison results (in \%) of the existing TZSL approaches with multi modalities on AwA and CUB datasets. Two types of semantic embedding space are used. $\mathcal{A}$ and $\mathcal{W}$ are short for attributes and word vector, respectively. $\mathcal{V}$ and $\mathcal{G}$ are short for VGG and GoogleNet features, respectively. For each dataset, the best one with VGG features is marked with underline and the best one with GoogleNet features is marked in bold.}}
\begin{center}
\begin{tabular}{|l|c|c|c|c|c|}
\hline
Method & $\mathcal{F}$ & SS &AwA & CUB & Average\\
\hline\hline
\multirow{3}{*}{SJE \cite{Akata15cvpr}} & \multirow{3}{*}{$\mathcal{G}$} & $\mathcal{A}$ & 66.7 & 50.1 & 58.4\\
& & $\mathcal{W}$ & 51.2 & 28.4 & 39.8\\
& & $\mathcal{A}$+$\mathcal{W}$ & 73.5 & 51.0 & 62.3\\
\hline
\multirow{3}{*}{LatEm \cite{xian2016cvpr}} & \multirow{3}{*}{$\mathcal{G}$} & $\mathcal{A}$ & 72.5 & 45.6 & 59.1 \\
& & $\mathcal{W}$ & 52.3 & 33.1 & 42.7\\
& & $\mathcal{A}$+$\mathcal{W}$ & 76.1 & 47.4 & 61.8\\
\hline
\multirow{3}{*}{RKT \cite{wang2016aaai}} & \multirow{3}{*}{$\mathcal{V}$} & $\mathcal{A}$ & 76.0 & 39.6 & 57.8\\
& & $\mathcal{W}$ & \underline{76.4} & 25.6 & 51.0\\
& & $\mathcal{A}$+$\mathcal{W}$ & 82.4 & 46.2 & 64.3\\
\hline
\multirow{3}{*}{BiDiLEL \cite{wang2016ijcv}}  & \multirow{3}{*}{$\mathcal{V}$} & $\mathcal{A}$ & 78.3 & 48.6 & 63.5\\
& & $\mathcal{W}$ & 57.0 & 33.6 & 45.3\\
& & $\mathcal{A}$+$\mathcal{W}$ & 77.8 & 51.3 & 64.6\\
\hline
\multirow{3}{*}{LSE} & \multirow{3}{*}{$\mathcal{V}$} & $\mathcal{A}$ & \underline{81.9} & \underline{55.6} & \underline{68.8}\\
& & $\mathcal{W}$ & 74.9 & \underline{35.2} & \underline{55.1}\\
& & $\mathcal{A}$+$\mathcal{W}$ & \underline{83.2} & \underline{56.3} & \underline{69.8}\\
\hline
\multirow{3}{*}{LSE} & \multirow{3}{*}{$\mathcal{G}$} & $\mathcal{A}$ & \textbf{81.6} & \textbf{53.2} & \textbf{67.4}\\
& & $\mathcal{W}$ & \textbf{76.3} & \textbf{34.7} & \textbf{55.5}\\
& & $\mathcal{A}$+$\mathcal{W}$ & \textbf{84.5} & \textbf{54.3} & \textbf{69.4}\\
\hline
\multirow{3}{*}{LSE} & \multirow{3}{*}{$\mathcal{G}$+$\mathcal{V}$} & $\mathcal{A}$ & \textbf{84.4} & \textbf{53.9} & \textbf{69.2}\\
& & $\mathcal{W}$ & \textbf{78.0} & \textbf{35.2} & \textbf{56.6}\\
& & $\mathcal{A}$+$mathcal{W}$ & \textbf{86.1} & \textbf{54.8} & \textbf{70.5}\\
\hline
\end{tabular}
\end{center}
\end{table}

\begin{table*}
    \label{Table.5}
    \caption{\upshape Performance (\%) comparison with the state-of-the-art approaches on GZSL. The best performance is marked in bold under different scenarios.}
    \centering
    \begin{tabular}{|c|c|c|c|c|c|c|c|c|c|c|c|c|c|c|c|c|}
    \hline
    \multirow{2}{*}{Method} &\multicolumn{4}{c|}{AwA} &\multicolumn{4}{c|}{CUB} &\multicolumn{4}{c|}{aPY} &\multicolumn{4}{c|}{ImageNet}\\
    \cline{2-17}
     &U-U &S-S &U-T &S-T &U-U &S-S &U-T &S-T &U-U &S-S &U-T &S-T &U-U &S-S &U-T &S-T\\
    \hline
    \hline
    ESZSL \cite{Romera15icml} &76.8 & 87.9 &26.2 &87.6 &49.1 &66.4 &15.1 &65.2&47.3&75.6&24.6&70.3&25.7 &\textbf{94.9} &9.1&\textbf{94.8}\\
    SynC \cite{Chao2016eccv}  &73.4& 81.0 &0.4 &81.0 &\textbf{54.4} &\textbf{73.0}&13.2 &\textbf{72.0} &44.2 &72.9 &18.4&68.9&23.4&93.7 &7.2&93.6\\
    JEDM \cite{yu2017trans} &77.4&84.2&31.9 &82.6&48.4 &51.4 &12.3 &48.9 &49.6&\textbf{76.4} &35.5&71.6&25.9&93.2 &8.8&93.0\\
    SAE \cite{Kodirov17cvpr} &81.4&84.7 &35.5 &85.2 &46.2 &57.1 &27.4 &56.7&41.3&71.7 &29.7&66.6&25.6 &92.9 &8.4 &92.2\\
    MFMR \cite{xu2017cvpr} &76.6 &81.2 &33.2 &79.7 &46.2 &48.6 &12.5 &42.8&46.4&65.3 &31.3&54.3&21.6&89.2 &6.9&88.9\\
    LSE &\textbf{82.0} &\textbf{88.2}& \textbf{42.4} &\textbf{87.9} &53.2 &64.1 &\textbf{33.6} &62.1&\textbf{53.9}&75.7 &\textbf{51.2}&\textbf{74.2}&\textbf{27.4}&93.6 &\textbf{12.4}&93.3\\
    \hline
    \end{tabular}
\end{table*}

One major limitation of many existing ZSL approaches is that they mostly focus on two modalities, e.g., the visual and the attribute modalities. However, the semantic information hides in different modalities in the real world. Thus, it is typically expected to develop the capability with more than two modalities. A main advantage of LSE is that it can fuse the multimodal features into the framework. To this end, we evaluate our approach with multimodal features on AwA and CUB datasets, including fusing the class semantic features (e.g., attribute and word vector) as well as visual features (e.g., VggNet and GoogleNet features). Four related multimodal confusion approaches are selected for comparison. SJE \cite{Akata15cvpr} and LatEm \cite{xian2016cvpr} are two GoogleNet-based approaches, and RKT \cite{wang2016aaai} and BiDiLEL \cite{wang2016ijcv} are two VGG-based approaches.

TABLE \uppercase\expandafter{\romannumeral5} summarizes our comparison results with the competing approaches on AwA and CUB datasets. From the table, we have the following observations: 1) Our LSE approach achieves the best performance on both datasets with different modalities except for the result using VGG as visual features and word vectors as semantic embedding features. Specifically, the proposed LSE outperforms the second best method BiDiLEL \cite{wang2016ijcv} with 5.3\%, 9.8\%, and 5.2\%  with VGG visual features using attribute, wordvec, and attribute+wordvec as semantic embedding features, respectively. Besides, with GoogleNet visual features, the proposed LSE has 8.3\%, 13.3\%, and 7.6\% gains over the second best one, i.e., LatEm \cite{xian2016cvpr} with attribute, wordvec, and attribute+wordvec, respectively. 2) On both datasets, the results of different ZSL approaches with attributes are better than those with word vectors, indicating that the attribute information contains more transferring semantics than the existing word vector representations. 3) The classification results of all ZSL approaches by exploiting both the attribute and word vector are much better than those with a single one, which demonstrates that both the attributes and word vectors retrieve not only the common information but also the complementary features. 4) In contrast to those of the AwA dataset, the results on CUB with word vectors are obviously inferior to those with attributes. This may be due in part to the fact that the CUB is a fine-grained dataset of which differences between inter-classes are small, making the word vectors to contain less discriminative information. 5) Fusing VGG features and GoogleNet features can further improve the performance, which indicates that different visual features contain complementary information. 


\subsection{Results of GZSL}

\begin{figure*}[t]
\begin{center}
  \includegraphics[height=11.5cm,width=0.95\linewidth]{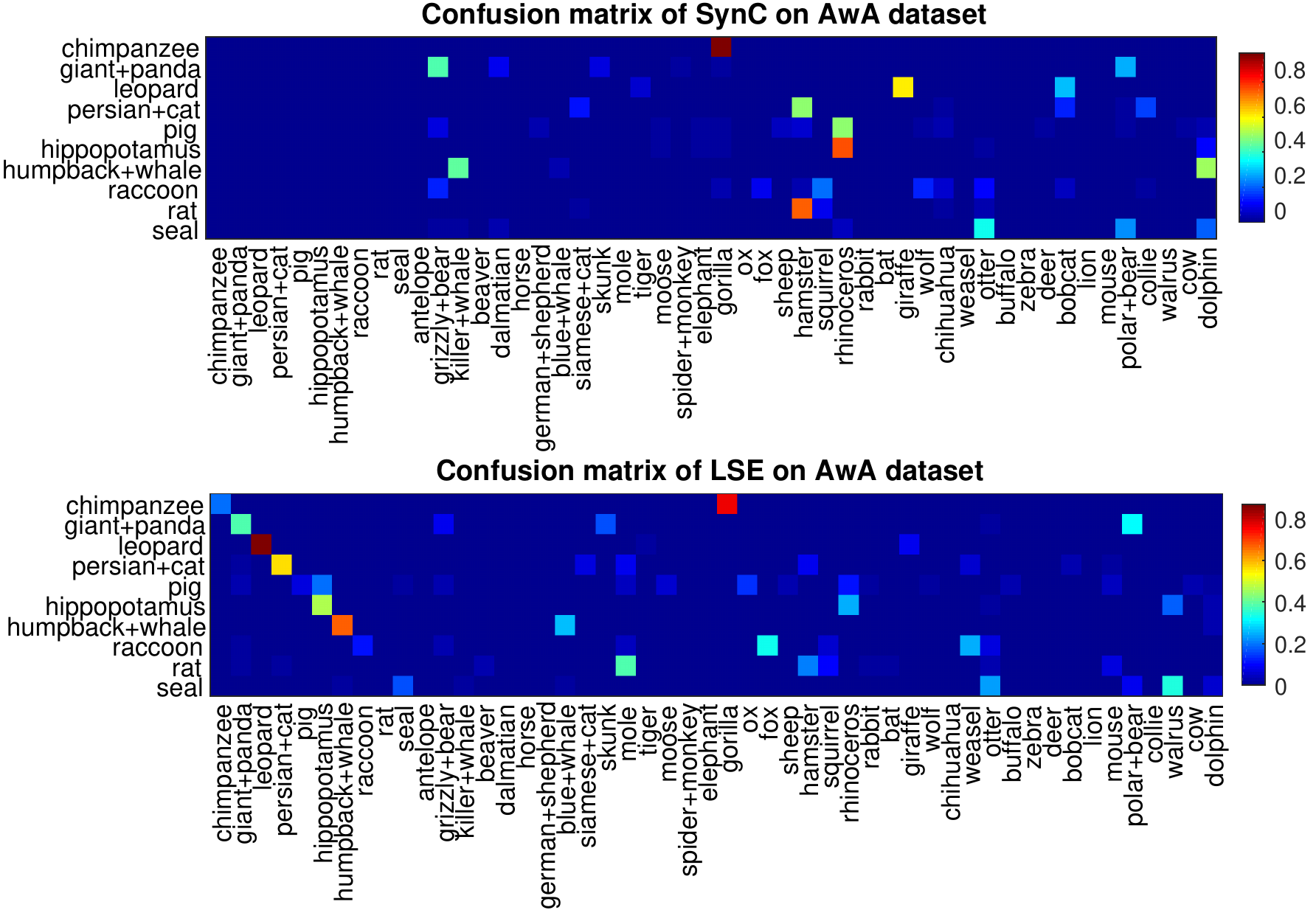}
\end{center}
   \caption{The confusion matrixes of SynC \cite{Chao2016eccv} and our LSE on AwA dataset under U-T scenario, where the columns are the classes that the testing instances truly belong to and the rows are the testing instances be classified into.}
\label{fig:fig1}
\end{figure*}

We also evaluate our approach to GZSL setting on the four datasets. Four scenarios U-U, S-S, U-T, and S-T are evaluated. U-U actually is TZSL, which means that the testing instances are assumed to be classified into the candidate unseen classes. S-S is the traditional supervised classification. In the experiments, 80\% instances from seen classes are randomly selected to train the model and the remaining 20\% instances are used to test. U-T is the scenario where the candidate classes of the testing instances from unseen classes are total classes, including both the seen and unseen classes, while S-T is to classify the testing instances from seen classes into both the seen and unseen classes. For these four scenarios, high performances of U-T and S-T are encouraged since they have more practical significance. TABLE \uppercase\expandafter{\romannumeral6} compares our model with five competitors on the four datasets. For AwA, CUB, and aPY datasets, per-class accuracy is used to evaluate the performance, while for the ImageNet dataset, top@5 classification accuracy is used. All the performances of the competitors are obtained via fine-tuning the models using the codes released by the authors. The hyperparameters of the competitors are selected from \{0.01,0.1,1,10,100\}.

 From the results shown in TABLE \uppercase\expandafter{\romannumeral6}, we observe that 1) the performance differences between S-S and S-T are small, which means that most testing instances from seen classes are classified into the seen classes. However, the performance differences between U-U and U-T are very large, which indicates that many testing unseen instances are wrongly classified into the seen classes. 2) LSE performs satisfactorily under U-T scenario and beats the other competitors by a large margin. Specifically, it has 6.9\%, 6.2\%, 15.7\%, and 3.3\% improvements over the second-best methods on AwA, CUB, aPY, and ImageNet datasets, respectively.

 In order to show a clearer comparison, we further visualize the classification results in terms of the confusion matrixes of SynC \cite{Chao2016eccv} and LSE on AwA dataset under U-T scenario. As illustrated in Fig.~2, SynC wrongly classifies most testing instances into the corresponding affinal seen classes. For example, the instances from ``chimpanzee" class, as an unseen class, are mostly classified into its affinal class ``gorilla", which is a seen class. However, for LSE, although many testing instances are also classified into the seen classes, most of them are classified into the correct classes. This indicates that the proposed approach not only enables transferring the information from the seen classes to unseen ones but also has the ability to distinguish the differences between seen classes and unseen ones. From the perspective of transferability, more affinal seen classes ensure the information to be easily transferred to the unseen classes. However, more affinal seen classes also alleviate the discriminability of the model since most unseen instances tend to be classified into the seen classes. The comparison results in the TABLE \uppercase\expandafter{\romannumeral6} and Fig.~2 illustrate that LSE finds a better tradeoff than the competing approaches.


\begin{table}
\caption{\upshape Zero-shot retrieval mAP (\%) comparison on three benchmark datasets. The results of the selected approaches are cited from the original papers.}
\begin{center}
\begin{tabular}{|l|c|c|c|c|}
\hline
Method & AwA &CUB & aPY & Average\\
\hline\hline
SSE-INT \cite{zhang15iccv} &46.3 & 4.7  &15.4 &22.1\\
JSLE \cite{Zhang16cvpr}   &66.5 & 23.9 &32.7 &41.0\\
SynC$^{struct}$ \cite{Changpinyo2016cvpr}   &65.4 & 34.3 &30.4 &43.4\\
MLZS \cite{bucher2016eccv}    &68.1 & 25.3 &36.9 &43.4\\
MFMR \cite{xu2017cvpr}    &70.8 & 30.6 &\textbf{45.6} &49.0\\
LSE     &\textbf{73.2} & \textbf{44.8} &42.3 &\textbf{53.4}\\
\hline
\end{tabular}
\end{center}
\end{table}

\subsection{The performances of ZSR }

Given an unseen class prototype as a query, the task of ZSR is to retrieve its related instances from unseen candidate set. In the experiment, the VGG visual features are used to obtain the retrieval performance. Five state-of-the-art VGG-based approaches are selected for comparison. Since no comparative ZSR approaches are evaluated in literature on ImageNet dataset, we conduct experiments on the rest three attribute datasets. TABLE \uppercase\expandafter{\romannumeral7} presents the ZSR results in terms of mAP. From the results, we can find that LSE performs the best on AwA and CUB datasets. Specifically, LSE obtains 2.4\% and 10.5\% mAP score gains over the best counterparts on AwA and CUB datasets. Furthermore, the proposed LSE significantly and consistently outperforms the closest competitor, i.e., MFMR \cite{xu2017cvpr}, by 4.4\% on average. The superior ZSR performances of LSE indicate that the strong visual similarity between the same corresponding classes of the different modalities and the effective semantic alignment across different modalities with our LSE approach.

\subsection{Parameter Sensitivity Analysis}

\begin{figure}[t]
   \centering
    \hspace{0.1in}
    \subfigure[]
    {
        \label{fig:fig3(a)}
\includegraphics[height=2.45cm,width=1.51in]{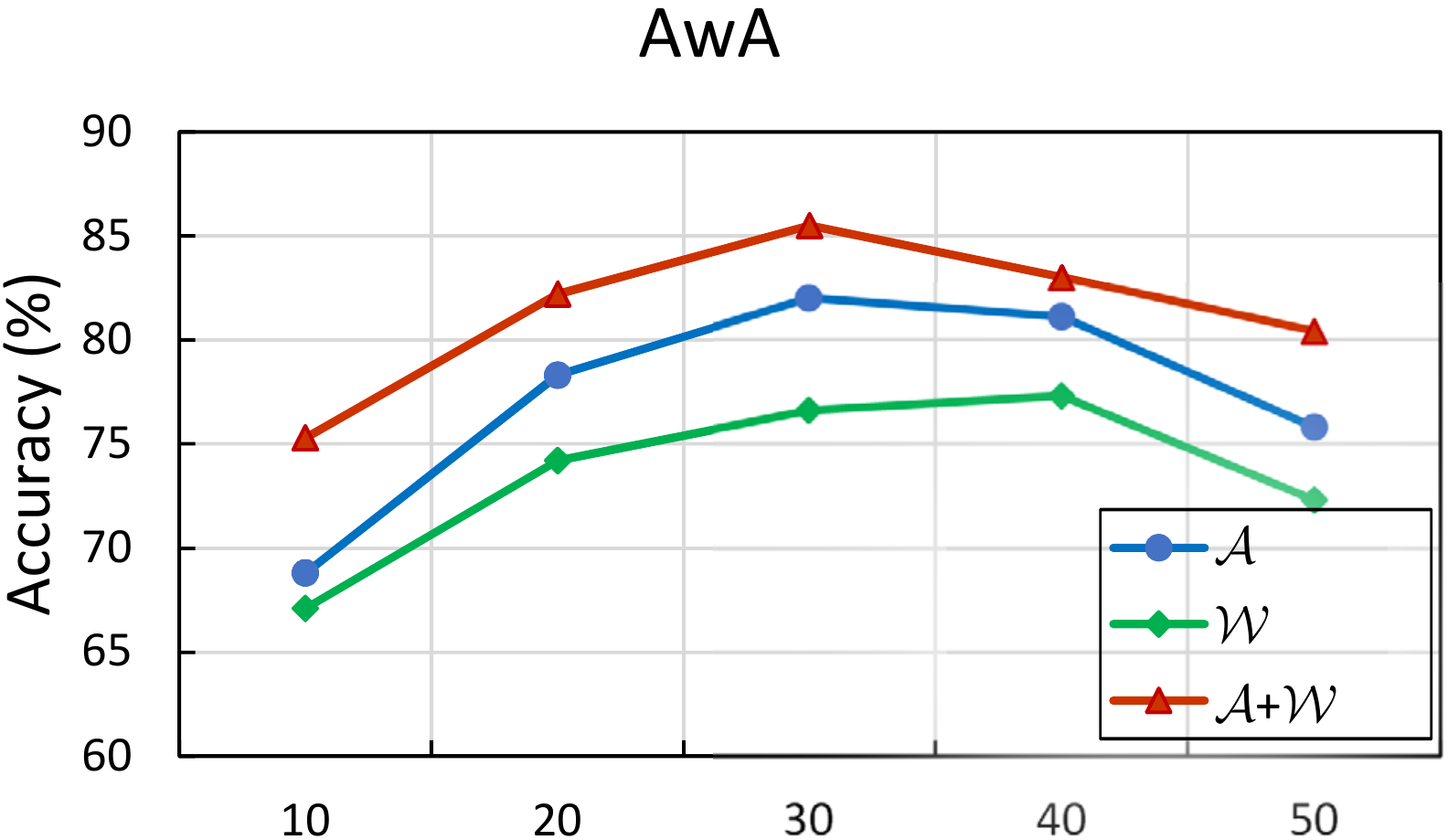}}
    \hspace{0.1in}
     \subfigure[]
    {
        \label{fig:fig3(b)}
\includegraphics[height=2.4cm,width=1.51in]{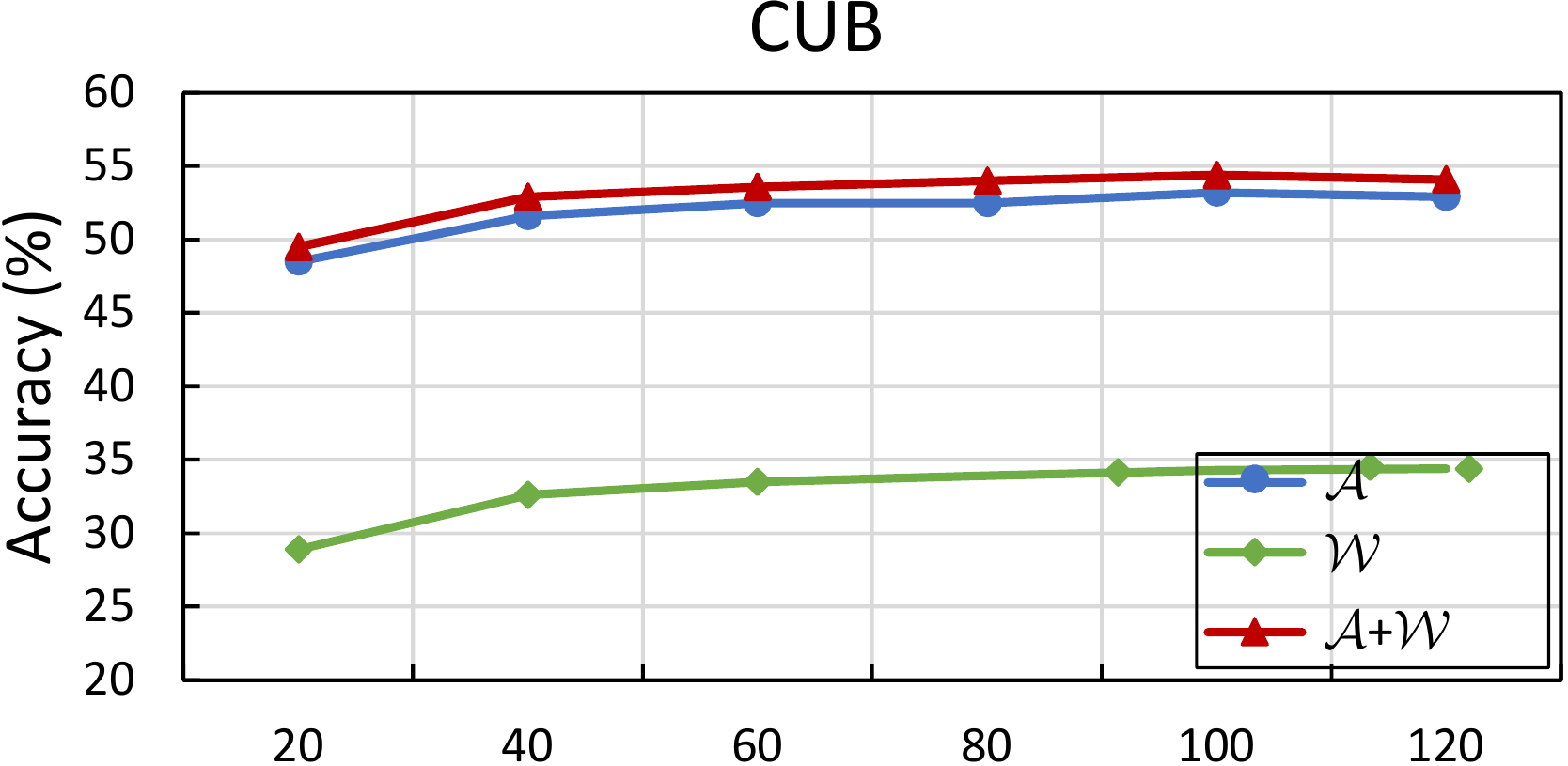}}
    \caption{\upshape{The influences of different latent dimensionalities $d$ on AwA and CUB datasets; $\mathcal{A}$ and $\mathcal{W}$ are short for attributes and word vector, respectively.}}
    \label{fig:fig3}
\end{figure}

\begin{figure}[t]
   \centering
    \hspace{0.1in}
    \subfigure[]
    {
        \label{fig:fig4(a)}
\includegraphics[height=2.35cm,width=1.5in]{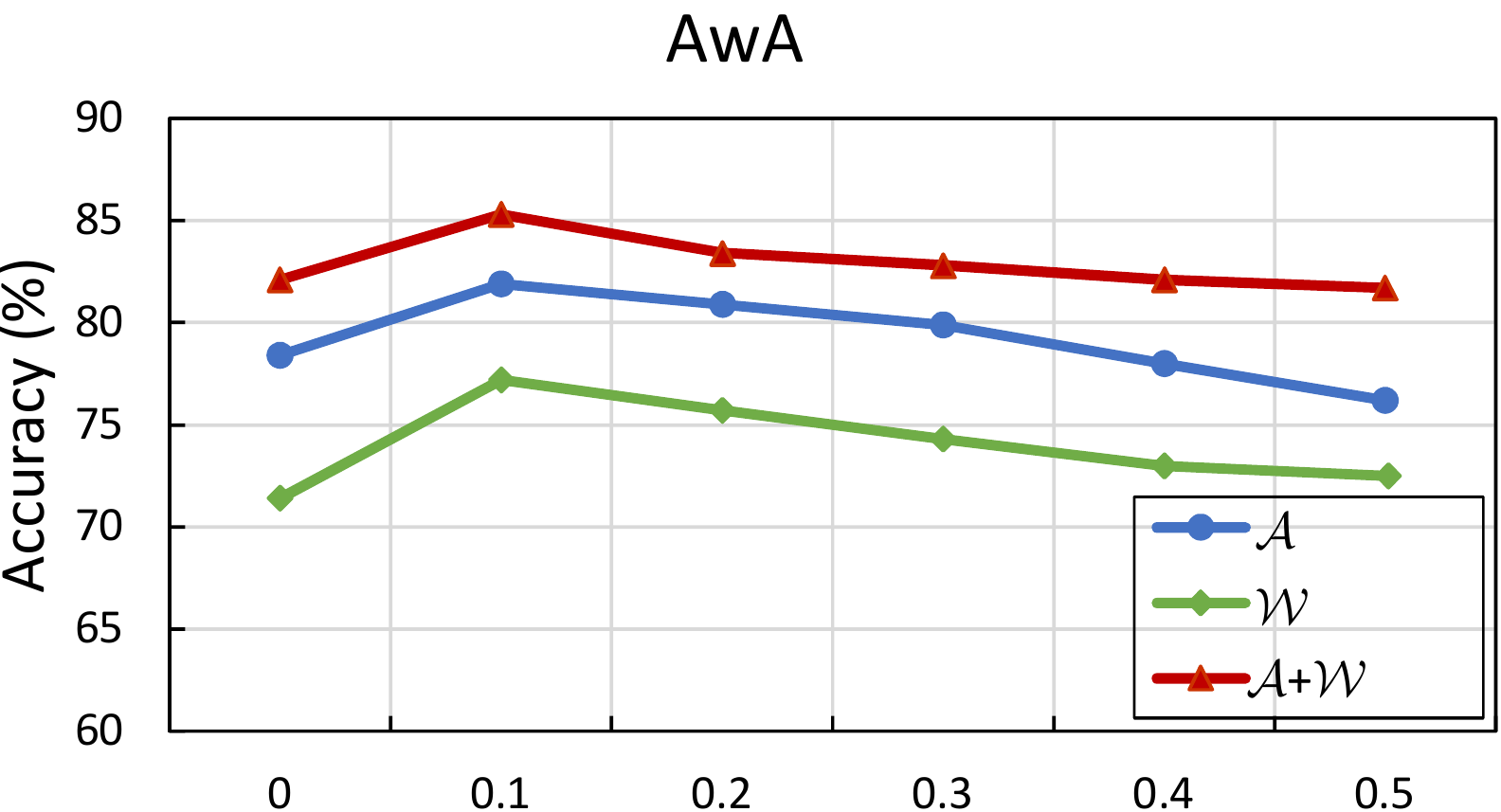}}
    \hspace{0.1in}
    \subfigure[]
    {
        \label{fig:fig4(b)}
\includegraphics[height=2.35cm,width=1.5in]{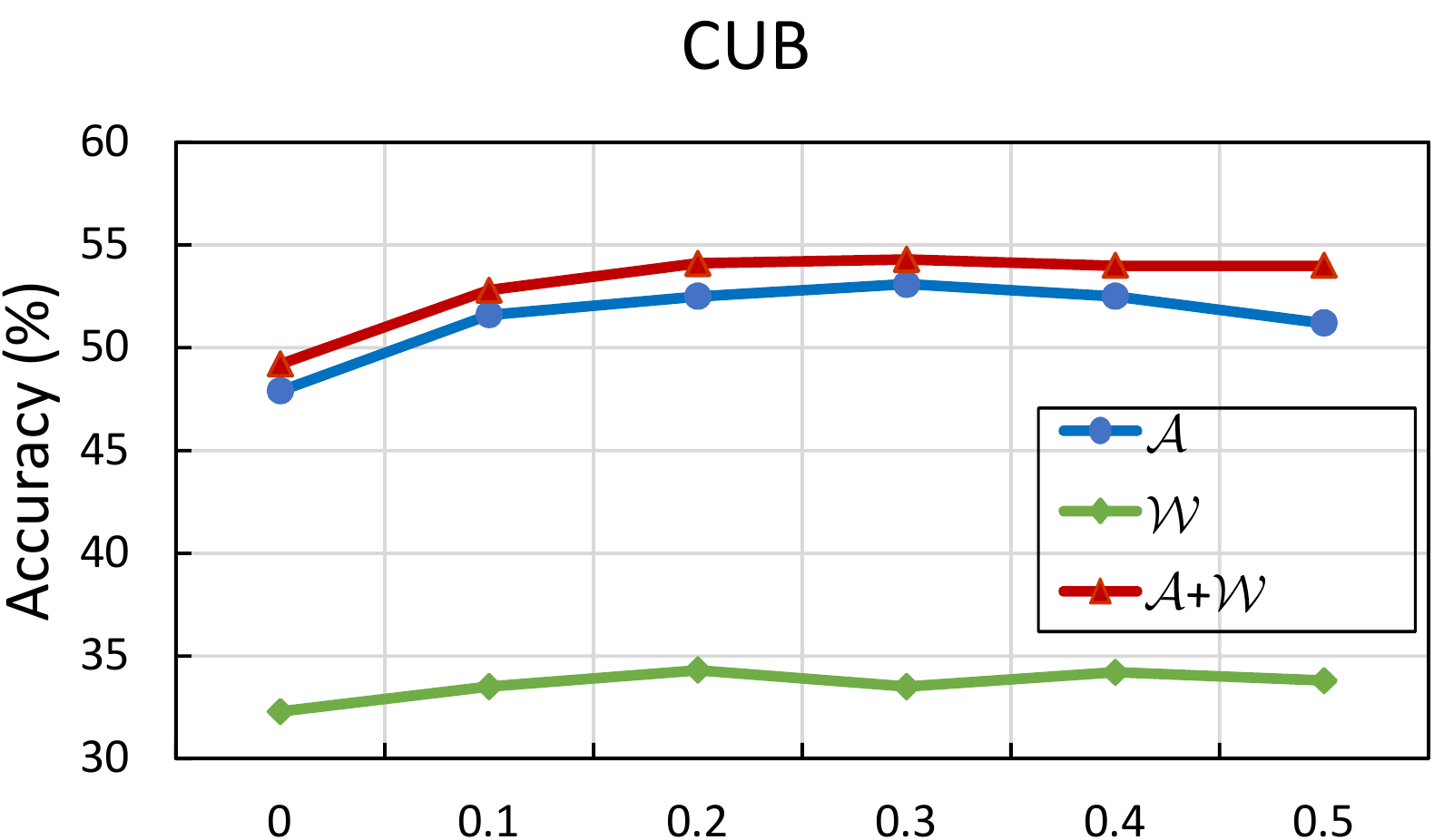}}
    \caption{The influences of different $\lambda$ on AwA and CUB datasets; $\mathcal{A}$ and $\mathcal{W}$ are short for attributes and word vector, respectively. }
    \label{fig:fig4}
\end{figure}
To evaluate the effects of the parameters in our method on unseen data, we take the TZSL classification accuracy on AwA and CUB datasets under different settings with respect to different parameter values. Specifically, there are two parameters $\lambda$ and $d$ in the training stage. $\lambda$ is the balance parameter, which is selected from $\textrm{[}0,1)$. $d$ is the dimensionality of the latent space, which is smaller than that of the input space. In the experiments, we vary one parameter at each time while fixing the other to its optimal value.

The two sub-figures in Fig.~3 illustrate the influences of different latent dimensionalities on AwA and CUB datasets. We observe that the curves vary on different datasets. This is reasonable since the dimensionalities of the original input semantic embedding spaces vary across different datasets. The curves in Fig.~3(a) show that the performances initially increase and achieve their peaks and then decline with the further increase of the latent dimensionality. We report the best performances on their peaks. However, the performances are more robust to the latent dimensionality on CUB dataset. As illustrated in Fig.~3(b), the curves tend to be flat when the latent dimensionality is larger than 60. The curves of these two datasets are drawn by setting balancing parameter $\lambda$ as 0.1.

Fig.~4 reports the classification performances using various $\lambda$ on AwA and CUB datasets. The two sub-figures of Fig.~4 show that, when $\lambda$ equals 0, the performances are worse than those with $\lambda$ equalling 0.1. This indicates that the encoding constraint (see Eq.~(\ref{Equ:equ1})) boosts the classification performances and improves the generalized transfer ability of the framework on unseen data. When $\lambda$ equals 0, the proposed approach turns to finding a latent space for both of different input modalities with matrix factorization, which has the same idea with MFMR \cite{xu2017cvpr}. As shown in Fig.~4(a), when $\lambda$ is larger than 0.1, the curves generally decrease with the increase of $\lambda$ on AwA dataset with different features, which indicates that the decoding process plays a more important role than the encoding process in the framework. The curves of CUB dataset in Fig.~4(b) have a similar trend to those in Fig.~4(a) but are more robust to the various values of $\lambda$. Compared to AwA dataset, the CUB dataset is a fine-grained dataset in which the differences of inter-classes are small. Thus, CUB is insensitive to $\lambda$. In the experiments, we set $\lambda$ as 0.1 for AwA and 0.2 for CUB datasets under different settings.

\subsection{Computational cost}

\begin{table}\label{tab:tab1}
\caption{\upshape{The computational cost (in second) of different linear ZSL approaches on AwA dataset.}}
\begin{center}
\begin{tabular}{|l|c|c|}
\hline
Method  & Training &Testing\\
\hline\hline
ESZSL \cite{Romera15icml} &0.62 & 0.04 \\
MFMC \cite{xu2017cvpr}       &66.95 & 1.01 \\
SynC$^{struct}$ \cite{Changpinyo2016cvpr} & 9.86 &4.22 \\
SAE \cite{Kodirov17cvpr} &1.19 & 0.34 \\
\hline\hline
LSE       & 57.76 & 0.42 \\
fast-LSE  & 0.60   & 0.42 \\
\hline
\end{tabular}
\end{center}
\end{table}

 \begin{table}
    \label{Table.5}
    \caption{\upshape  Performance (\%) comparison of LSE and fast-LSE on AwA and CUB datasets. $\mathcal{A}$ and $\mathcal{W}$ are short for attributes and word vector, respectively.}
    \centering
    \begin{tabular}{|c|c|c|c|c|}
    \hline
    \multirow{2}{*}{Method} &\multicolumn{2}{c|}{AwA} &\multicolumn{2}{c|}{CUB}\\
    \cline{2-5}
    & $\mathcal{A}$ & $\mathcal{W}$ &$\mathcal{A}$ &$\mathcal{W}$\\
    \hline
    LSE &81.6  &77.3 &53.2 &34.7 \\
    fast-LSE & 81.5 &76.8 &50.9  &33.8\\
    \hline
    \end{tabular}
\end{table}

In this section, we compare the computational cost of LSE with those of the other four linear-based ZSL approaches, ESZSL \cite{Romera15icml}, MFMC \cite{xu2017cvpr}, SynC$^{struct}$ \cite{Changpinyo2016cvpr}, and SAE \cite{Kodirov17cvpr}. As illustrated in TABLE \uppercase\expandafter{\romannumeral8}, we observe that LSE is a bit inefficient to the counterparts in the training stage. This is because the computational cost of LSE mainly comes from the eigenvalue decomposition in Eq.~(9), which depends on the number of training instances. As stated in \cite{Yu17tnn} that the transferability of a model depends on the representation of the training classes rather than the amount of the training instances. In this way, we apply LSE with the fast strategy proposed in \cite{Yu17tnn} by representing each training class as its visual pattern by averaging the visual features of the images in each class and call it as fast-LSE. The comparison results of LSE and fast-LSE are shown in TABLE \uppercase\expandafter{\romannumeral9}.
The results in TABLE \uppercase\expandafter{\romannumeral8} and TABLE \uppercase\expandafter{\romannumeral9} show that fast-LSE basically holds the classification performances, however, it improves the computational efficiency dramatically.
\section{Conclusion and future work}
 In this paper, we have proposed a novel latent space embedding approach for addressing ZSL. It learns the optimal intrinsic semantic information of different modalities via implicitly decomposing the input features based on an encoder-decoder framework. The proposed framework can also be extended to address the multimodal issues. Experimental results on TZSL, GZSL, and ZSR demonstrate that the proposed approach not only transfers the information from the seen domain to the unseen domain efficiently but also distinguishes the seen classes and unseen ones well.

 In the future, we will extend ZSL on some other related fields, such as saliency detection \cite{han18spm,zhang17pami,yao17tip}, person re-identification \cite{zeng17tii,liu15iccv}, and medical image analysis \cite{zhu17cvpr,yao16springer}. Besides, ZSL may be modified under weak supervision. The weak information could be accumulated in a self-paced way, just like \cite{zhang17cvpr,zhang16ijcai}. We will also explore ZSL under weakly supervision to progressively  incorporate training instances from easy to hard.

\appendices

\ifCLASSOPTIONcaptionsoff
  \newpage
\fi

%








\end{document}